# A First Approach on Modelling Staff Proactiveness in Retail Simulation Models

Peer-Olaf Siebers, Uwe Aickelin
School of Computer Science, University of Nottingham, Nottingham, NG8 1BB, UK

## Abstract

There has been a noticeable shift in the relative composition of the industry in the developed countries in recent years; manufacturing is decreasing while the service sector is becoming more important. However, currently most simulation models for investigating service systems are still built in the same way as manufacturing simulation models, using a process-oriented world view, i.e. they model the flow of passive entities through a system. These kinds of models allow studying aspects of operational management but are not well suited for studying the dynamics that appear in service systems due to human behaviour. For these kinds of studies we require tools that allow modelling the system and entities using an object-oriented world view, where intelligent objects serve as abstract "actors" that are goal directed and can behave proactively.

In our work we combine process-oriented discrete event simulation modelling and object-oriented agent based simulation modelling to investigate the impact of people management practices on retail productivity. In this paper, we reveal in a series of experiments what impact considering proactivity can have on the output accuracy of simulation models of human centric systems. The model and data we use for this investigation are based on a case study in a UK department store. We show that considering proactivity positively influences the validity of these kinds of models and therefore allows analysts to make better recommendations regarding strategies to apply people management practises.

## Introduction

To understand the impact of management practices on company performance is vital for a company's survival (Bloom et al. 2005). Empirical evidence affirms that both management practices and productivity measures must be context specific to be effective. Management practices need to be tailored to the particular organisation and the working environment, whereas productivity indices must also reflect a particular organisation's activities on a local level to be a valid indicator of performance (Siebers et al. 2008).

Often models for studying the impact of management practices on company performance are developed under the assumption of stability, equilibrium and linearity, whereas company's operations are considered in reality to be dynamic, non-linear and complex (Larsson and Petersson 2007). Furthermore, most of these models can only be applied as analytical tools once management practices have been implemented, i.e. they are not very useful at revealing system-level effects prior to the introduction of a specific management practice. This is most restricting when the focal interest is the development of the system over time, as dynamic behaviour is a fundamental feature of many interesting real-world phenomena. Alternatively, stochastic dynamic system simulation offers good potential to overcome such limitations. In particular Discrete Event Simulation (DES) modelling and Agent-Based Simulation (ABS) modelling are well suited for representing and analysing the dynamics of heterogeneous equilibrium and non-equilibrium systems and allow testing out different management strategies prior to their implementation.

There has been a lot of modelling and simulation of operational management practices, but people management practices, for example training, empowerment and teamwork, have often been neglected. This seems a fertile area for research as findings suggest that people management practices significantly impact on a business' productivity (e.g. Patterson et al. 1997; Rudman 2008; Birdi et al. 2008). One reason for the paucity of modelling and simulation of people management practices relates to their key component, an organisation's people, who are often unpredictable in their individual behaviour.

DES is a well established simulation approach in Operational Research (OR) and has been used for more than 40 years across a range of industries, amongst them manufacturing, travel, finance, and health (Hollocks 2006). Traditional DES is process oriented and therefore well suited for simulating workflow systems and investigating operational management practices. However, when the human element and its often unpredictable behaviour becomes a relevant factor for the analysis of system behaviour, as it is often the case in people centric system, traditional DES reaches its limits. People centric systems are systems that involve many human interactions and where the actors work with some degree of autonomy. Often some social aspects have to be considered as well.

We are investigating the impact of people management practices on company performance in the retail sector. In particular we are looking at the customer-staff interactions in department stores, which can be regarded as people centric systems, where system restrictions regarding the process flow are limited. Therefore, traditional DES does not seem to be the right tool for the investigations. In ABS models however, a system is modelled "bottom up" as a collection of autonomous and proactive decision making entities. Each entity individually assesses its situation and makes decisions on the basis of a set of adaptable rules (Bonabeau 2002). This modelling approach lends itself particularly well for modelling people and their behaviour. It allows considering autonomy and proactiveness and is used in related fields such as Social Science (e.g. Gilbert and Troitzsch 2005) and Economy (e.g. Tesfatsion and Judd 2006) to populate stochastic simulation models with entities that represent peoples' behaviour. However, these models are often very abstract, theoretical oriented and seek predominately qualitative answers while OR simulation models tend to be less abstract, more practical oriented, and seek often quantitative answers.

Therefore, in an earlier paper (Siebers et al. 2010a) we proposed using a combined DES/ABS modelling approach to design models that allow investigating people management practices. To demonstrate the usefulness of such an approach we conducted a case study with a top ten UK retailer to empirically inform the modelling and simulation process looking at Audio & TV (A&TV) and Womenswear (WW) departments across two branches. In these models we used a traditional DES modelling approach to represent the workflow of the system while we used an ABS modelling approach to represent human behaviour and social aspects. We demonstrated that this combination of DES and ABS modelling approach is a useful aid for studying the impact of people management practices on company performance. However, during our study we identified a shortcoming in our modelling approach which we are investigating further in this paper.

At first we used the staff rotas that we collected from the real system to define the number of staff and their roles (cashiers, normal service staff, or expert staff) in the simulation model. However, our validation experiments showed that with these settings it was impossible to get even close to the performance values observed in the real system. We then used an "idealised" staff rota. We kept the overall number of staff the same but changed the roles of some of the staff. In particular till manning was underrepresented in the original rota. However, the need for this "quick fix" indicates that we are not modelling the true dynamics and operational procedures of the real system. While we are focusing on reactive staff behaviour in our simulation model staff in the real system are encouraged and do act proactively, for example when there is a shortage of a particular staff type, staff members swap roles to help out. This is sometimes co-ordinated by a section manager but most of the time initiated by the staff themselves.

In this paper, we investigate the impact that incorporating proactive staff behaviour into our simulation model has on simulation output accuracy. We want to find out if this allows us to better represent the true dynamics and operational procedures of the real system and therefore improves the usability and credibility of the simulation model. From a practitioner's point of view we want to discover if the extra effort of adding proactivity to the simulation model is worth the effort, i.e. does it produce a significant improvement of simulation output accuracy that justifies the extra costs for data collection and modelling.

In our latest simulation model we have integrated some forms of proactive staff behaviour. We have then tested how different levels of staff proactiveness impact on the performance output of the simulation model (i.e. service experience, sales, and staff utilisation). Eventually we were able to use the real rota data and receive the correct performance values as they appeared in the real system. Therefore, we can have much more confidence in the experimental results when we test new management practices for the real world system. In addition, this result underpins the importance of using a combined DES/ABS modelling approach rather than a traditional DES modelling approach when studying people centric systems.

## Background

The development of the Agent-Based Modelling and Simulation (ABMS) technology has been enormous in the last couple of years; however real world empirical based applications found in Business Management are still very sparse. This is despite the recognition of the usefulness of simulation as a decision support tool for business analysts and managers.

### *ABMS and Business Management*

ABMS, with its intrinsic multidisciplinary approach, is gaining increasing attention as a problem-solving tool in the social sciences, particularly in economics, business, finance, and politics (Consiglio 2007) and is also on the advance in OR (Bonabeau 2002). It is a bottom-up modelling approach where aggregate dynamics emerge from the interactions of constituent components and between components and the environment (Gilbert and Terna 2000). It provides a powerful tool to study the dynamics of equilibrium and non equilibrium systems over time and its outputs offer the potential to be used for explanatory, exploratory and predictive purposes (Twomey and Cadman 2002). Examples for business management case studies can be found in Bonabeau (2002), Twomey and Cadman (2002), and North and Macal (2007).

### *Modelling people related management practices in OR*

We have found that most of the work relevant to our investigations focuses on marketing and consumer behaviour rather than on management practices. For example, Said et al. (2002) have created an ABS model composed of a virtual consumer population to study the effects of marketing strategies in a competing market context. A similar approach has been used by Baxter et al. (2003) who have developed an intelligent customer relationship management tool using ABMS that considers the communication of customer experiences between members of a social network, incorporating the powerful influence of word of mouth on the adoption of products and services. A very different facet of consumer behaviour is presented by Kitazawa and Batty (2004) and Casti (1999). Both study the retail movements of shoppers but at a different scale. While the first paper is concerned with movement in shopping centres the second focuses on movement in an individual supermarket. There are many more examples where ABMS has been employed to study consumer behaviours (e.g. Cao 1999; Csik 2003; Jager et al. 2000; Baydar 2003) or entire consumer market behaviours (e.g. Vriend 1995; Twomey and Cadman 2002; Said and Bouron 2001; Koritarov 2004; Janssen and Jager 2001; Schenk et al. 2007).

While most of the relevant papers we found do apply ABMS, there are some noteworthy exceptions. For example, Berman and Larson (2004) use queue control models to investigate the efficiency of cross-trained workers in stores. Another interesting contribution is made by Nicholson et al. (2002), who compare different marketing strategies for multi channel (physical and electronic) retailing, applying a traditional Belkian analysis of situational variables in a longitudinal study of consumer channel selection decisions. As a last example, we want to mention Patel and Schlijper (2004) who use a multitude of different analytical and other modelling approaches to test specific hypotheses of consumer behaviour.

We have also found one of-the-shelf software named ShopSim (2010), which is a decision support tool for retail and shopping centre management. It evaluates the shop mix attractiveness and pedestrian friendliness of a shopping centre. The software uses an agent-based approach, where the behaviour of agents depends on poll data.

### *Modelling proactive service behaviour in OR*

As we have stated in the introduction of this paper we want to focus on modelling some proactive staff behaviour at an operational level. The OR literature does not provide any guidance on how to implement such proactive behaviour into a simulation model. In fact we did not find any papers that explicitly discuss the topic of proactive service behaviour modelling and simulation. Therefore we have been looking in the related fields of management and agents. In the management literature we can find some definitions of what proactive service behaviour means at an operational level while in the agent literature we can find some indications of what proactivity means in technical terms. By relating these definitions to our specific case of department store staff proactiveness we then come up with a strategy of modelling our proactive staff behaviour at an operational level.

First, we consider the management literature to understand what proactivity means in relation to service provision. Rank et al. (2007) define *proactive customer service* as a self started, long term oriented, and persistent service behaviour that goes beyond explicitly prescribed requirements and *proactive service performance* as being related to an employee's trait personal initiative, affective organisational commitment, self efficacy, task complexity, task autonomy, and their supervisor's participative as well as transformational leadership. Several researchers, amongst them Crant (2000) and Frese and Fay (2001) have emphasised that proactive behaviour is different from prescribed behaviour, because the former involves long term oriented courses of self-started actions rather than mere responses to demand. Second, we consider how the agent community defines proactivity. The agent literature distinguishes between *reactive strategies* and *proactive strategies*. A reactive strategy triggers a response to actions as they occur while a proactive strategy triggers deliberate actions in anticipation of future problems (Hudson 2000). Proactive behaviour of agents is often modelled in terms of goals that the agents pursue (Van Riemsdijk et al. 2008). Goals have two aspects: declarative (a description of the state sought) and procedural (a set of plans for achieving the goal) (Winikoff et al. 2002). Now, by relating these definitions to our specific case, we define that our staff agents' (declarative) goal is to provide best service by proactively (self started and long term oriented) balancing the different queues that appear in a department store.

The importance of short waiting times for customers when requiring assistance or at the till has been emphasised in many papers, amongst them Bennett (1998) and Simmons (1989). In particular the till operations need to be running smoothly at any point in time as the till is the place where finally the money is made and it is disastrous to lose a customer at the last moment. However, other queues need to be observed as well to achieve high service quality. Well timed proactive behaviour of service staff can make a big difference to customer service satisfaction and therefore the success of the store.

In our case study department staff is encouraged to act proactively and we believe that this has a big impact at the operational level and is therefore important to be modelled. In order to achieve the goal of proactively balancing the different queues we have decided to model a set of plans that allows normal service staff to be proactively helping out as cashiers if they see the need arising. There are different indicators to start (e.g. till queue length above critical mark and help queue length below critical mark; relation between those two queues) and to finish this activity (help queue length above critical mark; till queue length below critical mark). More details are presented later in the paper in the "Model Design and Implementation" section under "Implementation".

## Model Design and Implementation

### Our approach

When modelling people management practices in OR one is mainly interested in the relations between staff and customers but it is equally important to consider parts of the operational structure of the system in order to enable a realistic evaluation of system performance and service experience. Therefore, we have decided to use a combined DES/ABS modelling approach. We model the system itself using a DES modelling approach while we model the people within the system using an ABS modelling approach (Siebers et al. 2010b). A queuing system will be used for representing the operational structure of the system while the people within the system will be modelled as a heterogeneous population of autonomous entities (agents) in order to account for the stochasticity caused by human decision making. This will also allow us to consider the long term effects of service quality on the decision making processes of customers, an important component for designing a tool for studying people centric system. The application of a combined DES/ABS modelling approach has been proven to be quite successful in areas like manufacturing and supply chain modelling (e.g. Parunak et al. 1998; Darley et al. 2004) as well as in the area of modelling crowd movement (e.g. Dubiel and Tsimhoni 2005).

When modelling people inside a system it is important to consider that there are differences in the way ABMS is applied in different research fields regarding their empirical embeddednes. Boero and Squazzoni (2005) distinguish between three different levels, amongst others characterised by the level of empirical data used for input modelling and model validation: case-based models (for studying specific circumscribed empirical phenomena), typifications (for studying specific classes of empirical phenomena) and theoretical abstractions (pure theoretical models). While case based models use empirical data for input modelling as well as model validation theoretical abstractions use no empirical data at all. Social Science simulation applications tend to be more oriented towards the bottom end of this scale (theoretical abstractions) OR applications are usually located at the top end (case based). This implies that there is also a difference in knowledge representation and in the outcome that the researcher is interested in. In Social Sciences it is common to model the decision making process itself (e.g. Rao and Georgeff 1995) and the focus of attention on the output side is on the emergence of patterns. On the other hand in OR applications the decision making process is often represented through probabilities or simple if-then decision rules collected from an existing real system (e.g. Schwaiger and Stahmer 2003; Darley et al. 2004) and the focus on the output side is on system performance rather than on emergent phenomena (e.g. detection of behavioural patterns). As we are studying a people centric service system, we have added some additional measures to assess how people perceive the services provided, besides the standard system performance measures.

*Choice of case study systems*

To test the applicability of using a combined DES/ABS modelling approach for studying the impact of people management practices on retail performance we conducted a case study. This case study involved four departments across two branches of a leading UK retailer. We have chosen two very different types of departments in order to make sure that our results regarding the application of people management practices are applicable for a wide variety of departments. We collected our data in the A&TV and the WW department of the two case study branches, which differ substantially, not only in their way of operating but also their customer base and staffing setup.

The two departments can be characterised as follows. In the A&TV department the average customer service times is much longer, the average purchase is much more expensive, the likelihood of customers seeking help is much higher, the likelihood of customers making a purchase after receiving help is lower, the conversion rate (likelihood of customers making a purchase) is lower, and the department tends to attract more solution demanders and service seekers. The terminology is explained later in this section under "Implementation". In the WW department the average customer service times is much shorter, the average purchase is much less expensive, the likelihood of customers seeking help is much lower, the likelihood of customers making a purchase after receiving is much higher, the conversion rate is higher, and the department tends to attract shopping enthusiasts

*Model design*

During our conceptualisation process we have gathered information about the departments and their operations, developed some agent templates and created the environment in which the agents exist and interact. Needless to say that these steps have not been consecutive and that there have been many iteration loops during this phase.

The case study work involved extensive data collection techniques, spanning: participant observation, semi-structured interviews with team members, management and personnel, completion of survey questionnaires on the effectiveness of retail management practices and the analysis of company data and reports. Research findings were consolidated and fed back (via report and presentation) to employees with extensive experience and knowledge of the four departments in order to validate our understanding and conclusions. This approach has enabled us to acquire a valid and reliable understanding of how the real system operates, revealing insights into the working of the system as well as the behaviour of and interactions between the different actors within it.

During our data collection phase we have collected two kind of data; global data about the processes embedded in the system (e.g. staffing levels, queue lengths, and customer mix) and individual oriented data about the behaviour of people (e.g. the different states customers and staff can be in, the time customers and staff spend in these states, and the likelihood of decisions).

For the conceptual design of our agents we have used state charts. Such a state chart shows the different states an entity (in our case a person) can be in and defines the events that cause a transition from one state to another. This is exactly the information we need in order to represent our agents at a later stage within the chosen simulation environment.

As an example Figure 1 represents the state chart for customer agents. Boxes represent states and arrows transitions. The customer state chart has an "entering" and a "leaving" state which symbolise a customer agent entering or leaving the store. In the centre is the "contemplating" state, a dummy state that links all other states. From the dummy state all customer activities can be initiated (browsing, queuing at till, seeking help, and seeking refund). While some activities (e.g. browsing) require only one step to be completed others (e.g. paying) require several steps to be completed successfully. However, all of the multi-step activities can be interrupted and the customer can go back to the contemplating state to start with another action or leave the store. Transitions happen either after a time-out (e.g. "browsing") or when certain conditions are met (e.g. leave "queuing for help" when staff available). The transition rules have been omitted here to keep the chart comprehensible. They are explained in more detail later in this section under "Implementation".

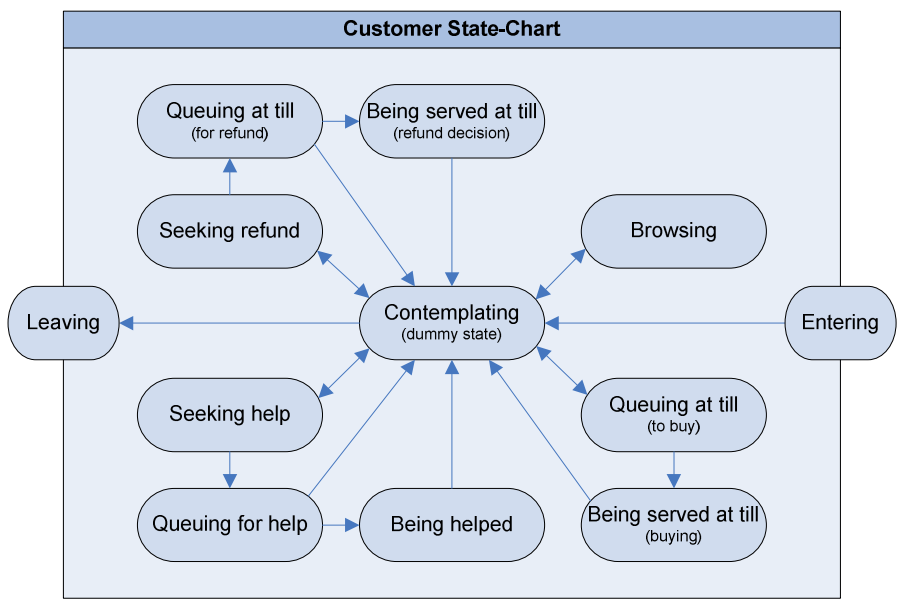

Figure 1: Conceptual state chart for a customer agent

We have found this graphical representation also very useful for the agent concept validation process because it is easier for an expert in the real system (who is not an expert in ABMS) to quickly take on board the model conceptualisation and provide useful validation of the model component structures and content.

An important remark about the order in which a customer is going through the different states within the customer state chart is that there is a logical order to events. Some of this order has been expressed with single- and double-headed arrows whereas others would have been difficult to express in the graph directly without losing the concept of the connecting contemplating state. For example, a customer would not be queuing at the till to buy something without having browsed for an item first. Therefore, the preceding event for a customer queuing at the till is to browse and then pick up the item s/he wishes to purchase. These rules have been considered later in the implementation.

There are two different sets of performance measures that should be considered for evaluating the performance of people centric systems: service performance and productivity/profitability. While there are many direct objective measures for evaluating productivity and profitability, there are no direct objective measures for evaluating service performance, as this is a question of how the individuals perceive the service given and provided (Parasuraman et al. 1988). Therefore, in general service performance measures are based on subjective opinions of customers and staff, and difficult to be produced as an output of a simulation model.

A measure that provides some information about service performance is service experience. Here we propose a novel way of estimating service experience based on the waiting times for each individual service requested during a single visit. Furthermore, customer type and previous shopping experiences influence the time people are prepared to wait and therefore the chances of leaving the store with a positive shopping experience, even if it is crowded and people have to queue for a substantial amount of time.

In order to be able to estimate service experience we introduce a service level index using satisfaction weights. This allows a customer's service experience to be recorded throughout the simulated lifetime of the customer. The idea is that certain situations might have a bigger impact on service experience than others, and therefore weights can be assigned to events to account for this. Applied in conjunction with an ABMS approach, we expect to observe interactions with individual customer differences; variations which have been empirically linked to differences in service experience (e.g. Simon and Usunier 2007). This helps the analyst to find out to what extent customers underwent a positive or negative shopping experience and it also allows the analyst to put emphasis on different operational aspects and try out the impact of different strategies.

As an example, Figure 2 shows the implementation of the "help" loop of the customer state chart. Here, rounded rectangles represent states, arrows transitions, circles with a B inside branches, and numbers represent the satisfaction weights. Assuming that a customer needs help, s/he would transit into a "seek help" state. In this state s/he will be looking for service staff. After some time has passed s/he has either found a staff member that can help (positive experience, satisfaction score) or has to wait for customer service as no staff member is available (negative experience, satisfaction score). In this case the customer has to join the queue. In particular in peak times service queues can be quite long in service intensive department like A&TV. For some customers the waiting time might be too long and they leave the queue after a certain amount of time. This situation often results in the perception of having wasted a lot of time and not received any service, which is a very negative experience and therefore this situation is linked to a high negative satisfaction score. This customer would leave the "help" loop with a satisfaction score of -2 -4 = -6. If someone is served immediately, the overall satisfaction score for the "help" loop would be +2 +2 = 4. Someone who has to wait but receives the required service in the end will leave wit a neutral score of -2 +2 = 0. The same scoring systems but with different satisfaction weights are in place for all the other loops ("pay" and "refund") of the customer state chart. The final satisfaction score for a department store visit is calculated by summing up all the individual scores from that particular visit.

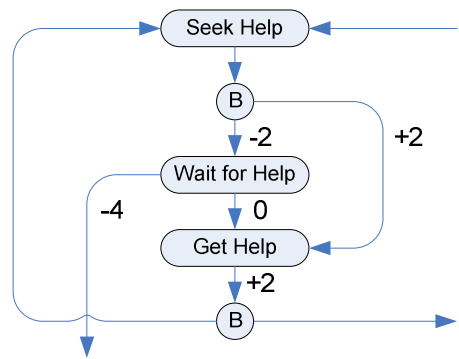

Figure 2: The "help" loop implementation of the customer agent state chart

Each customer agent stores two different satisfaction scores. The first score (accumulated history) considers the shopping history of the agent and adds or subtracts the absolute score from the current shopping experience to the historic score calculated from all the previous experiences. The second score (experience per visit) records if the current shopping experience has been positive, neutral or negative, and counts the overall number of positive, neutral and negative experiences.

The values we are using for the satisfaction weights are based on a customer survey we have conducted in our case study departments. We asked the customers how important prompt service is to them for the different services they would request (help, pay, or refund) and how a delay would influence their satisfaction. The survey results indicate that customers are more patient when they have to wait for pay and refund compared to when they have to wait for help. Therefore, the values of the satisfaction weight are higher in the "help" loop compared to the "pay" and "refund" loop.

A key aspect to consider is that the most interesting system outcomes evolve over time and many of the goals of the retail company (e.g., service standards) are planned strategically over the long-term. Consumer decision making, and consequently changes in the behaviour of customers over time, are driven by an integration of each consumer's cognitive and affective skills (Hansen 2005). At any given point in time a customer's behaviour, as the product of an individual's cognitions, emotions, and attitudes, may be attributable to an external social cue such as a friend's recommendation or in-store stimuli (Tai and Fung 1997), or an internal cue such as memory of one's own previous shopping experiences (Youn and Faber 2000).

We use a finite population of customer agents (customer pool) to study the perception of service and customer behaviour changes due to changes in service quality. In our model we currently only consider internal influences. Each of our customer agents is equipped with a long-term memory and can adapt its preferences and behaviours over time through internal stimulation triggered by memory of one's own previous shopping experiences.

So how does it all fit together? An overview of our conceptual department store model including all types of entities and their interactions is presented in Figure 3. To explain the figure and clarify how the different entities interact with each other and how the overall system works we now present a step-by-step walk-through.

In the beginning of each simulation run a new customer pool is created that usually contains a large population of heterogeneous customer agents which are equipped with a long term memory influencing their decision making process. The size of the customer pool can be defined through a setup parameter before the execution. After being created the customers remain in a resting state. Also, a staff pool is created that contains the staff of different types and different levels of expertise according to the requirements of a weekly rota. The size of the staff pool depends on the highest number of staff required on an individual day. There can be different amounts of staff at week days and weekends. After being created the staff members also remain in a resting state. Throughout the simulated opening hours customers are activated at random with an inter-arrival time according to hourly customer arrival rates. While customers are picked dynamic throughout the day staff members are picked only once at the beginning of the day. Once picked, staff members go immediately in a waiting state, i.e. waiting to be addressed by a customer via a signal (asking for some kind of service).

Once a customer has entered the department s/he will be in the contemplating state. This is a dummy state and represents the reality of an individual thinking through their behavioural intentions prior to acting (Ajzen 1985), regardless of whether the department visit will result in a planned or unanticipated purchase (Kelly et al. 2000). Even when a particular purchase is planned, the consumer may change their mind and go for a substitute product, if they buy at all. S/he will probably start browsing and after a certain amount of time, s/he may require help, queue at the till or leave the shop. If the customer requires help, s/he considers what to do and seeks help by looking for a staff member (i.e. sending a signal to the staff members on duty for that day that s/he wants help) and will either immediately receive help if a staff member of the right type and right level of expertise is available (in waiting state) or wait for attention. If no staff member is available, s/he has to join a queue and wait for help. If the queue moves very slowly it could happen that s/he gets fed up waiting or runs out of time and leaves the queue prematurely (an autonomous decision). This does not mean necessarily that s/he will not make a purchase. Sometimes customers would still buy the item even without getting the advice they were seeking. Another reason why a customer might come into the department is to ask for a refund. From an organizational point of view the refund process is very similar to the help process. The difference is that the refund process will take place at the till. This might affect till waiting times, if authorisation by management is required but management is not available immediately. After the refund process is concluded the customer will either continue shopping (i.e. start browsing) or leave the department.

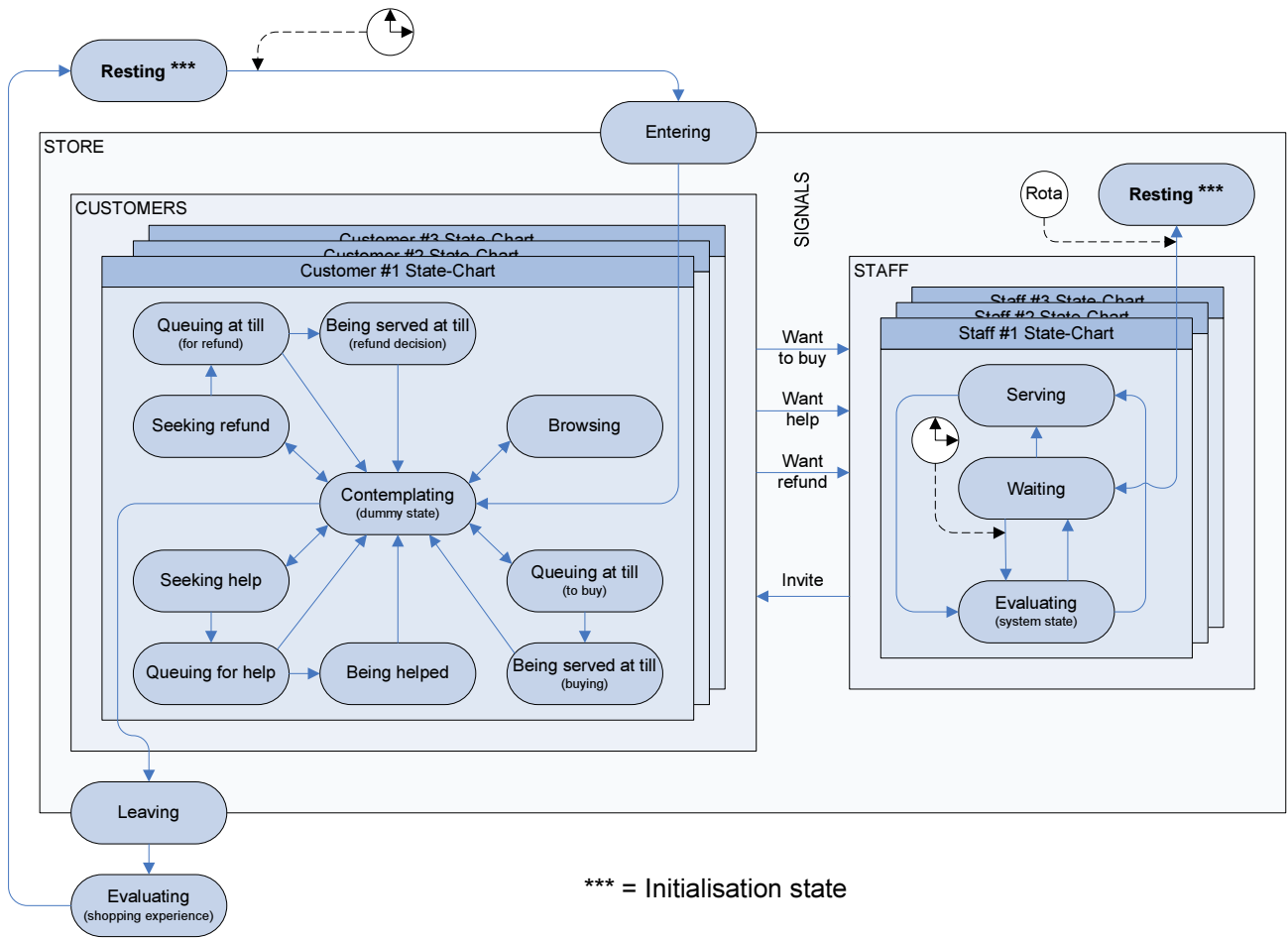

Figure 3: Conceptual department store model

Once a customer has finished shopping s/he will leave the department and evaluate her/his shopping experience. Whether the shopping experience has been positive or negative will influence the level of patience during the next visit. After this reflective form of behaviour a customer will go back into resting state until reanimated by being randomly picked for the next shopping trip.

Staff members can do more than just respond to the signals sent by customers. They can behave proactively by approaching customers and offering services (by sending an invite signal). This can for example happen when they notice that the queue at the till get very long or they might offer help in a role that is different to their assigned role (i.e. an expert staff can offer to help out as a normal service staff member or a section manager if there is a need for it). When a staff member has finished serving a customer and in defined intervals (based on observations we made in the case study departments) staff will evaluate the state of the system (around them) and respond if they see an opportunity to improve the customer experience by reducing waiting times either for customers seeking help, waiting to pay or waiting for refund. If a staff member decides proactively to open a new till it will stay open for a while, if expert staff help out as normal service staff it can be a one-off solution. This is close to what we observed in the real system where it would look bad to customers if a cashier would close a till even though the situation of the system (the queue length) remains beyond a reasonable length, while the number of staff providing advice is usually not observed actively, so after helping out once a staff member might focus again on his/her primary role.

*Implementation*

We have implemented our conceptual department store model in AnyLogic™ v5.5. AnyLogic is a Java™ based multi-paradigm simulation software (XJTEK 2005) that supports the development of combined DES and ABS models based on state charts and some java code. It allows replacing passive objects typical for DES models with active objects (or agents), which we have used to represent the actors of the real world system.

In this subsection we will give a brief overview of how we implemented the concepts described above. However, as the focus of this paper is proactive staff behaviour this will be the main topic here. Some details about other aspects of the implementation can be found elsewhere; the base model is described in Siebers et al. (2009) and the implementation of the customer pool is described in Siebers et al. (2010a).

First, we look at *knowledge representation*. Often agent logic is based on analytical models or heuristics and in the absence of adequate empirical data theoretical models are used. We have taken a different approach and used frequency distributions for determining state change delays and probability distributions for representing the decisions made as statistical distributions are the best format to represent the data we have gathered during our case study due to their numerical nature. In this way, we have created a population with individual differences between agents, mirroring the variability of attitudes and behaviours of their real human counterparts.

Our frequency distributions are modelled as triangular distributions supplying the time that an event lasts, using the minimum, mode and maximum duration. We have chosen triangular distributions here as we have only a relatively small sample of empirical data and a triangular distribution is commonly used as a first approximation for the real distribution (XJTEK 2005). The values for our triangular distributions are based on our own observation and expert estimates in the absence of numerical data. We have collected this information from the two branches and calculated an average value for each department type, creating one set of data for A&TV and one set for WW. Table 1 lists some sample frequency distributions that we have used for modelling the A&TV department (the values presented here are slightly amended to comply with confidentiality restrictions). The distributions are used as exit rules for most of the states. All remaining exit rules are based on queue development, i.e. the availability of staff.

| Situation | Min. | Mode | Max. |
|---|---|---|---|
| Leave browse state after … | 1 | 7 | 15 |
| Leave help state after … | 3 | 15 | 30 |
| Leave pay queue (no patience) after … | 5 | 12 | 20 |

Table 1: Sample frequency distribution values

The probability distributions are partly based on company data (e.g. conversion rates, i.e. the percentage of customers who buy something) and partly on informed guesses (e.g. patience of customers before they would leave a queue). As before, we have calculated average values for each department type. Some examples for probability distributions we used to model the A&TV department can be found in Table 2. The distributions make up most of the transition rules at the branches where decisions are made with what action to perceive (e.g. decision to seek help). The remaining decisions are based on the state of the environment (e.g. leaving the queue, if the queue does not get shorter quickly enough).

| Event | Probability of event |
|---|---|
| Someone makes a purchase after browsing | 0.37 |
| Someone requires help | 0.38 |
| Someone makes a purchase after getting help | 0.56 |

Table 2: Sample probabilities

The complete set of the frequency and probability distributions used in the simulation model is provided together with the source code (Siebers 2010). We have also gathered some company data about work team numbers and work team composition, varying opening hours and peak times, along with other operational and financial details (e.g. transaction numbers and values) some of which we have used as input data for our model (e.g. work team composition) and some of which we have used to validate our model against (e.g. transaction numbers).

Second, we look at the *implementation of customer types*. In real life, customers display certain shopping behaviours that can be categorised (Reynolds and Beatty 1999). Hence, we defined customer types to create a heterogeneous customer base, thereby allowing us to test customer populations closer to reality. We have introduced five different customer types: shopping enthusiasts, solution demanders, service seekers, disinterested shoppers and internet shoppers. The first three customer types have been identified by the case study organisation as the customers who make the biggest contribution to their business, in terms of both value and frequency of sales. In order to avoid over-inflating the amount of sales that we model we have introduced two additional customer types, which use services but often do not make a purchase (e.g. internet shoppers ask for advice in store, but buy on the internet). The definition of each customer type is based on the customer's likelihood to perform a certain action, classified as either: low, moderate, or high, as given in Table 3. With help of the department managers of the participating case study departments we were then able to define the typical customer population of the different department types. These are as follows: WW customers can best be represented as 80% shopping enthusiast and 20% disinterested shoppers while A&TV customers can best be represented as 5% shopping enthusiasts, 40% solution demanders, 40% service seekers, 5% disinterested shoppers and 10% internet shoppers.

| Customer type | Likelihood to | | | |
|---|---|---|---|---|
| | buy | wait | ask for help | ask for refund |
| Shopping enthusiast | high | moderate | moderate | low |
| Solution demander | high | low | low | low |
| Service seeker | moderate | high | high | low |
| Disinterested shopper | low | low | low | high |
| Internet shopper | low | high | high | low |

Table 3: Definitions for each type of customer.

We have developed two algorithms to imitate the influence of the customer type attributes (likelihood to perform a certain action) on customer behaviour. They have been implemented as methods that are invoked when defining the state change delays modelled by frequency distributions, and when supporting decision making modelled by probability distributions. Basically these methods define new threshold values for the distributions based on the likelihood values mentioned above. Figure 4 shows as an example the pseudo code for the probability distribution threshold correction algorithm. If the customer is a shopping enthusiast and is about to make the decision whether to make a purchase or to leave the department directly a corrected threshold value (probability) for this decision is calculated. For this calculation the original threshold of 0.37 (Table 2) is taken into account and, for a shopping enthusiast where there is a high likelihood to buy (Table 3), the corrected threshold value is calculated as follows: 0.37+0.37/2 = 0.56. Consequently the likelihood that a shopping enthusiast proceeds to the checkout rather than leaving the department without making any purchase has risen by 18.5%.

```
for (each threshold to be corrected) do  {
   if (OT < 0.5) limit = OT/2 else limit = (1-OT)/2
   if (likelihood = 0) CT = OT – limit
   if (likelihood = 1) CT = OT
   if (likelihood = 2) CT = OT + limit
}

where:   OT = original threshold
         CT = corrected threshold
         likelihood: 0 = low, 1 = moderate, 2 = high
```

Figure 4: Pseudo code for the probability distribution threshold correction algorithm

Third, we look at the *implementation of staff proactiveness*. During our case study we were told by the department managers that staff are encouraged to act proactively and we also observed proactive behaviour on the shop floor. For example when there is a shortage of a particular staff type staff members would swap roles to help out.

While the traditional DES approach only supports the modelling of reactive behaviour a combined DES/ABS modelling approach supports the modelling of reactive and proactive behaviour (as each agent has its own threat of execution). In this paper we want to demonstrate the benefits of this capability and we have therefore implemented some of the proactive work routines that we have observed in the real system.

As argued earlier in the "Background" section under "Modelling proactive service behaviour in OR" the most important proactive staff behaviour is that non-cashier staff does open and close tills proactively, depending on demand (i.e. length of till queues) and staff availability. However, there are several rules in form of task priorities that need to be considered before non-cashier staff members can either start or stop helping out as a temporary cashier. These are:

- Priority 1: continue as temporary cashier unless a stop strategy (see below) has come true
- Priority 2: if expert staff, help out as section manager (might be required for refund process)
- Priority 3: if normal service staff or expert staff, help out as temporary cashier
- Priority 4: if expert staff, help out as normal service staff
- If none of these is applicable, wait for a given time and then check again if role swap is required

The pseudo code for implementing this proactive staff behaviour is presented in Figure 5. There are three different stop strategies that allow temporary cashiers to swap back to their standard roles. A parameter that can be defined in the model setup defines which strategy is to be used. Using the first strategy a temporary cashier swaps back once he has served a predefined number of customers. Using the second strategy a temporary cashier swaps back once the queue has reached a predefined critical length. The third strategy is a combination of both, where a temporary cashier swaps back once he has served a predefined number of customers or once the queue has reached a predefined critical length, whatever happens first.

```
if staff is a temporary cashier {
  if one of the active proactive stop strategies has come true {
    // close down checkout
    change current staff type (cashier) to original staff type (expert or normal) and maintain staff counters
  }
}
if staff is expert {
  if expert is required to help out as section manager {
    // help out as section manager
    invite customer from section manager queue
    exit sub
  }
}
if normal service or expert staff is required to help out as temporary cashier {
  if there is enough staff to cope with their original task {
    if the maximal number of tills is not already open {
      if the queue at each check out is of critical length {
        // open new checkout
        change original staff type (expert or normal) to staff type cashier and maintain staff counters
      }
    }
  }
}
// serve in original staff type (expert, normal, cashier) or new assigned one (cashier)
invite customers from the appropriate queue (expert, normal, or cashier)
if successful, exit sub
if expert staff {
  // help out at normal service staff queue
  invite customers from normal service staff queue
  if successful, exit sub
}
If no temporary cashier is required (cashier queue empty) {
  // close down checkout
  change current staff type (cashier) to original staff type (expert or normal) and maintain staff counters
  // check again if released staff is required to work as section manager or in his/her own staff type role
  if staff is expert {
    if expert is required to help out as section manager {
```

```
    // help out as section manager
    invite customer from section manager queue
    exit sub
   }
  }
 }
 invite customers from the appropriate queue (expert, normal, or cashier)
 if successful, exit sub
 if expert staff {
  // help out at normal service staff queue
  invite customers from normal service staff queue
  if successful, exit sub
 }
}
```

Figure 5: Pseudo code for proactive staff behaviour implementation

As we want to analyse the impact of different proactive staff strategies on system performance and service experience in our experiments, we have parameterised the setting for proactive staff involvement so that they can be defined in the model setup. Overall we have six parameters. The first two define the values for the stop strategies, the third defines the minimum number of staff required to cope with the standard task of a potential temporary cashier and therefore controls the maximum number of staff of a certain type that can swap at the same time, the fourth defines the maximum number of tills physically available, the fifth defines which of the three stop strategy is to be used for the experiment, and the sixth defines the mean time between two checks if a role swap is required. The default settings for these six "proactivity" parameters are as follows:

- P1: Maximum number of customers to serve as a temporary cashier: 10
- P2: Critical queue length for opening/closing additional tills: 3 (opening); 2 (closing)
- P3: Minimum number of staff required to cope with original task: 2
- P4: Maximum numbers of open tills: 4 (A&TV); 6 (WW)
- P5: Stop strategy: Stop service as temporary cashier when P1 or P2 has been reached
- P6: Check if support at the till is needed every 2 minutes (random checks are switched off!)

Finally, we look at the *implementation of new performance measures*. Besides the service performance measures we have described earlier in this section under "Model design" we have added two utilisation performance measures and some statistics. These help us to understand whether the staff team's composition is effectively meeting the demands placed on it by customers but also give us an insight into the level of proactivity occurring in the simulated system. While the first performance measure collects the staff utilisation (for studying the effectiveness of staff team composition) the second measure looks at busy times of staff in their different roles (for studying the level of proactivity). Finally we collect some statistics on how often staff are changing their roles and how long they stay in these non-standard roles. Utilisation performance measures are recorded as weekly averages.

The staff utilisation measure collects statistics on what staff are employed as (the staff type as defined by the rota) and not what they are actually doing (staff role as defined by proactivity), i.e. even if normal staff are helping out as cashiers the busy time will be counted towards the normal staff utilisation measure. This is the standard way of collecting utilisation data. It is calculated by taking the average busy times of the staff members of a specific staff group (cashiers, normal service staff, and expert staff) and dividing this value by the sum of the idle and busy times of the staff members of this specific staff group (considering only the working hours).

The busy times by role measure collects some information about how much time the staff members spend in the different roles (as cashiers, normal service staff, and expert staff). Together with the statistics on how often staff are changing their roles and how long they stay in these non-standard roles it provides a measure for the level of proactivity initiated by the different settings of the six "proactivity" parameters mentioned above.

In order to make the simulation model outputs more meaningful for managers we have added some monetary performance measures. As such measures are context-specific and vary by sector we have made our selection of productivity and profitability measures to be considered by looking at the appropriate literature (e.g. Sellers-Rubio and Mas-Ruiz 2007; Crespi et al. 2006; Griffith and Harmgart 2005). Our chosen monetary performance measures include: overall staff cost per day, sales turnover, sales per employee, sales per employee hour worked, net profit per employee, and net profit per hour worked.

## Verification and Validation

While verification is conducted only during the model development phase validation is a process that is conducted throughout a simulation study life cycle (from defining the objectives of the study right through to implementing the proposed changes). For our verification and validation process we have used a framework proposed by Robinson (2004). Although this framework has been designed for DES studies and we conduct a combined DES/ABS study it is applicable as our agents are embedded in an operational structure and therefore the overall system behaviour (macro behaviour) does not emerge merely from the interaction of the micro entities as it is the case in a genuine ABS model (Pourdehnad 2002). In our case, the system behaviour primarily stems from the flow of the entities through the system which in our case is influenced by the reactive and proactive behaviour of the entities itself. Therefore we are able to validate our model qualitatively and quantitatively at the micro and macro level.

The verification and validation process proposed by Robinson (2004) comprises conceptual model validation, data validation, white box (micro) validation, black box (macro) validation, experiment validation, solution validation, and verification. First, to ensure that the content, assumptions and simplifications of our conceptual model are sufficiently accurate for the purpose at hand we have discussed our state charts, pseudo code, assumptions and simplifications with some of the employees of the case study departments that have extensive experience and knowledge of the operational and behavioural aspects of the case study system. At the same time, we have asked them to take a look at the collected data and confirm that they are in line with what they would have expected. In addition by conducting some cross-checks we made sure that our secondary data is reliable. After implementing our conceptual model we thoroughly verified our simulation model by checking the source code for bugs and by checking that all algorithms have been implemented properly. Next we conducted white-box validation, where we compared the content of the simulation model (at the micro level) to the real world system, to make sure that our agents represent their real world counterparts in sufficient accuracy for the purpose at hand and on the other hand that all operational processes (queuing, customer/staff engagement, etc.) represent the real world processes in sufficient accuracy for the purpose at hand.

As we stated in the introduction of the paper when we performed our black-box validation (macro check of the simulation model's operation) with the previous version of this model we encountered some problems with model validity and we used a "quick fix" to get simulation model outputs that match with the performance data we collected from the case study departments. We have now introduced a new mechanism (staff proactiveness) to overcome these validity problems and it is the purpose of the experiments in the next section to repeat the black-box analysis to confirm that our new mechanism has solved the problem and therefore eliminated the need for a "quick fix". In the next section we also do some experiment validation (sensitivity analysis) to gain some insight into the sensitivity of the parameters that we use to control the level and form of proactiveness that we want to investigate.

With regards to the validity of our experimental parameters (run length, number of replications) which is part of the experiment validation, we can safely use the values that we determined for the previous version of this simulation model (Siebers et al. 2010a) as our experimental framework has not changed. Finally, solution validation does not apply here, as we have as yet not investigated any specific managerial problems for the client and therefore no solution or recommendations have so far been implemented.

## Experimentation

There are different reasons for model execution: validation and sensitivity analysis (summarised as validation experiments) and system operation experiments and system optimisation experiments (summarised as system experiments). While the first two are concerned with model operation (comparing the model to the real system behaviour) the latter two are concerned with system operation (working on a fictive system). In this paper we present a series of validation experiments we have conducted to test the new features of our simulation model. While the first experiment is concerned with model validity the second experiment is a calibration exercise and a sensitivity analysis in one. A copy of the simulation model configured with the standard set-up values that we used for the experimentation is available online (Siebers 2010).

During the experimentation we used the same set-up values for our simulation model input parameters (one set for the A&TV department and one for the WW department) and varied only the values of the experimental variables. Department-specific parameters include things like staffing level, service times, and the composition of the customer pool. During simulation execution we use average values for opening hours and footfall. Each of the simulation runs simulates a period of 52 weeks. Although on a first view the system looks like a terminating system one needs to keep in mind that the entities are equipped with a long term memory and therefore we need to simulate more than just a single day per simulation run. We conducted 20 replications for each model configuration, using the confidence interval method with a 95% confidence interval proposed by Robinson (2004) to ensure that the number of replications would be sufficient. In fact, the required number of replications turned out to be much lower (only two), but as the simulation runs very quickly we decided to conduct more runs to reduce the variance of the simulation outputs.

*Experiment 1: Empirical validation*

Our first experiment is a validation experiment. We conduct a comparison study, testing the model with and without our new staff proactiveness features and with different staffing levels. Furthermore, we test our new performance measures. In broader terms, we want to learn how the dynamics of the simulation model have changed due to the addition of the staff proactiveness features, if our new utilisation performance measures will be able to give us some useful information to better understand these dynamics, and finally, if our new productivity and profitability performance measures present some information that is useful for management decision making.

In this experiment we use two different sets of staffing levels, "optimised staffing levels" and "real staffing levels". The phrase "optimised staffing levels" refers to what we thought would be the staffing levels that best represent the effects of proactive staff behaviour that we observed during our case study data collection, i.e. we considered how staff allocated their time in different roles. The phrase "real staffing levels" refers to the staffing level data we have collected during our case study from the original weekly rota (for the experiment we use the week day staffing levels only). Both staffing levels are shown in the "Parameter Settings" section of Table 4 (for A&TV) and Table 5 (for WW). Note, that the "optimised staffing levels" include a smaller number of staff compared to the "real staffing levels".

It is difficult to empirically evaluate agent-based models, in particular at the macro level, because the behaviour of the system emerges from the interactions between the individual entities (Moss and Edmonds 2005). However, as we are dealing with agents embedded in a process oriented simulation model (with a queuing structure), we expect our simulation model macro level output to reflect the performance we have measured in the real system.

As we have not collected data on shopping experience in the case study departments (which would have been our first choice) we have decided to use the weekly number of transactions as our key performance measure for assessing the accuracy of the simulation model output in comparison to the real system performance. In our analysis we conduct a quantitative comparison of the real system transaction data (average of four weeks) with the simulation model results produced by employing different simulation model settings. Then we use all the other output data that we are collecting from the simulation model to gain more insight into the effect that considering proactivity has on the simulation dynamics. In our analysis we use these for a qualitative comparison of the results produced by employing different simulation model settings.

Our research question for this experiment can be phrased as follows: *Can we get more accurate results from our simulation models (in particular number of transactions), now that we are modelling some proactive staff behaviour features directly in our simulation model, when using real staffing levels (i.e. our collected rota data)?*

To answer this question we have used the following experimental design:

- Scenario a: optimised staffing levels, proactiveness off
- Scenario b: optimised staffing levels, proactiveness on
- Scenario c: real staffing levels, proactiveness off
- Scenario d: real staffing levels, proactiveness on

In this context "proactiveness off" means staff is only serving customers that require their staff type, while "proactiveness on" means that expert staff is helping out as normal service staff and normal service staff is helping out at the till, however, their level of commitment is subject to the proactiveness parameter settings. We have conducted the experiment for both department types (A&TV and WW) and have used the default proactiveness parameter settings. We have not conducted a rigorous statistic analysis on the results as we only want to demonstrate the new features of the simulation model here. However, for the qualitative comparison we provide a point estimate (mean) as well as a measure of spread (standard deviation) to allow taking into account the variability inherent in the simulated system.

The results shown in Table 4 for A&TV and Table 5 for WW (number of transaction in the real world and simulated scenarios and deviation of the simulation results from the real world data) allow us to assess the improvement in empirical validity. Here is a quick reminder of our goal. It is to achieve similar or smaller deviations from the real system data in scenario d (real staffing) compared to scenario a (optimised staffing).

| Parameter Settings | Audio & TV (outputs: means of weekly averages; deviations: relate to the real world value) | | | | | | | | |
|---|---|---|---|---|---|---|---|---|---|
| Scenario | real world | a | | b | | c | | d | |
| Staffing | real | optimised | | optimised | | real | | real | |
| Number of {cashiers; normal staff; expert staff} | {1;10;1} | {2;6;2} | | {2;6;2} | | {1;10;1} | | {1;10;1} | |
| Proactive | yes | no | | yes | | no | | yes | |
| Outputs | Mean | Mean | Deviation | Mean | Deviation | Mean | Deviation | Mean | Deviation |
| Transactions | 1787.09 | 1203.43 | 32.66% | 1387.25 | 22.37% | 615.70 | 65.55% | 1346.22 | 24.67% |

Table 4: Weekly number of transactions for A&TV in the real world and the simulated scenarios

| Parameter Settings | Womenswear (outputs: means of weekly averages; deviations: relate to the real world value) | | | | | | | | |
|---|---|---|---|---|---|---|---|---|---|
| Scenario | real world | a | | b | | c | | d | |
| Staffing | real | optimised | | optimised | | real | | real | |
| Number of {cashiers; normal staff; expert staff} | {2;13;1} | {3;8;2} | | {3;8;2} | | {2;13;1} | | {2;13;1} | |
| Proactive | yes | no | | yes | | no | | yes | |
| Outputs | Mean | Mean | Deviation | Mean | Deviation | Mean | Deviation | Mean | Deviation |
| Transactions | 3172.35 | 2931.43 | 7.59% | 3138.00 | 1.08% | 2004.97 | 36.80% | 2977.20 | 6.15% |

Table 5: Weekly number of transactions for WW in the real world and the simulated scenarios

In the tables we can see that if we consider proactiveness in our simulation model we can improve the validity of the results for real staffing massively (scenario c compared to d) and get close to the results that we achieved with using optimised staffing (scenario a). This is true for both department types and gives us confidence that we are getting much better in representing the true dynamics of the real system. However, the mean transaction values for A&TV for scenario d (1346.22) are still significantly lower than the mean of the collected data (1787.09). This could be a consequence of using the wrong settings for our proactivity parameters. We will investigate this issue further in our second experiment.

Table 6 and 7 represent the performance outputs of the simulation model produced by employing the different simulation model settings (scenario a-d). Table 6 presents the non-monetary and monetary performance outputs for the A&TV department while Table 7 represents the non-monetary and monetary performance outputs for the WW department. We provide the output in form of mean value and standard deviation.

| Measure | Parameter Settings | Audio & TV | | | | | | | |
|---|---|---|---|---|---|---|---|---|---|
| | Scenario | a | | b | | c | | d | |
| | Staffing | optimised | | optimised | | real | | real | |
| | Number of {cashiers; normal staff; expert staff} | {2;6;2} | | {2;6;2} | | {1;10;1} | | {1;10;1} | |
| | Proactive | no | | yes | | no | | yes | |
| | Non-monetary performance measures (means of weekly averages; times provided in minutes) | | | | | | | | |
| | Outputs | Mean | SD | Mean | SD | Mean | SD | Mean | SD |
| 1 | Overall number of customers | 4086.07 | 9.42 | 4086.56 | 6.11 | 4086.37 | 8.47 | 4086.60 | 8.82 |
| 2 | Number of customers that leave happy (purchase) | 1203.43 | 1.53 | 1387.25 | 3.00 | 615.70 | 1.41 | 1346.22 | 4.24 |
| 3 | % of customers that leave happy (purchase) | 29.45% | 0.08% | 33.95% | 0.07% | 15.07% | 0.04% | 32.94% | 0.08% |
| 4 | % of customers that leave not waiting for normal help | 2.93% | 0.07% | 2.87% | 0.07% | 0.13% | 0.01% | 0.45% | 0.02% |
| 5 | % of customers that leave not waiting for expert help | 0.93% | 0.02% | 1.04% | 0.02% | 1.80% | 0.03% | 1.91% | 0.03% |
| 6 | % of customers that leave not waiting to pay | 17.12% | 0.12% | 12.57% | 0.07% | 32.68% | 0.12% | 14.54% | 0.06% |
| 7 | % of customers left satisfied >0 (accumulated history) | 79.69% | 0.29% | 81.80% | 0.33% | 74.89% | 0.28% | 88.28% | 0.21% |
| 8 | % of customers left neutral =0 (accumulated history) | 5.83% | 0.06% | 6.17% | 0.09% | 5.22% | 0.10% | 4.59% | 0.06% |
| 9 | % of customers left unsatisfied <0 (accumulated history) | 14.48% | 0.28% | 12.03% | 0.29% | 19.89% | 0.25% | 7.13% | 0.20% |
| 10 | % of customers left satisfied >0 (experience per visit) | 40.31% | 0.14% | 41.04% | 0.13% | 38.68% | 0.08% | 46.99% | 0.13% |
| 11 | % of customers left neutral =0 (experience per visit) | 44.83% | 0.11% | 46.91% | 0.12% | 40.89% | 0.12% | 42.47% | 0.13% |
| 12 | % of customers left unsatisfied <0 (experience per visit) | 14.86% | 0.09% | 12.05% | 0.10% | 20.43% | 0.09% | 10.54% | 0.06% |
| 13 | Average time customers spend in department (per visit) | 20.72 | 0.03 | 20.84 | 0.03 | 20.54 | 0.03 | 20.68 | 0.03 |
| 14 | Average time customers spend waiting for help (per visit) | 5.74 | 0.04 | 5.33 | 0.04 | 3.57 | 0.03 | 3.85 | 0.03 |
| 15 | ... as % of time customers spend in help block | 29.09% | 0.20% | 26.37% | 0.14% | 19.29% | 0.17% | 20.34% | 0.15% |
| 16 | % of customers that have to wait for help | 55.58% | 0.48% | 69.82% | 0.52% | 21.50% | 0.30% | 36.92% | 0.59% |
| 17 | Average time customers spend waiting to pay (per visit) | 9.20 | 0.02 | 8.57 | 0.01 | 11.13 | 0.01 | 9.01 | 0.01 |
| 18 | ... as % of time customers spend in pay block | 73.21% | 0.10% | 68.56% | 0.06% | 86.57% | 0.04% | 69.91% | 0.03% |
| 19 | % of customers that have to wait to pay | 97.63% | 0.06% | 97.63% | 0.08% | 99.45% | 0.01% | 99.43% | 0.01% |
| 20 | Utilisation: cashiers | 97.45% | 0.08% | 97.50% | 0.10% | 99.66% | 0.05% | 99.64% | 0.07% |
| 21 | Utilisation: normal staff (excluding proactive support) | 83.72% | 0.22% | 85.50% | 0.19% | 63.99% | 0.20% | 74.17% | 0.22% |
| 22 | Utilisation: expert staff (excluding proactive support) | 78.62% | 0.42% | 85.84% | 0.39% | 68.14% | 0.34% | 74.99% | 0.32% |
| 23 | Utilisation: all staff (including proactive support) | 85.44% | 0.08% | 87.97% | 0.08% | 67.31% | 0.19% | 76.36% | 0.20% |
| | Monetary performance measures (means of daily averages) | | | | | | | | |
| | Outputs | Mean | SD | Mean | SD | Mean | SD | Mean | SD |
| 24 | Overall staff cost per day [£] | 7680.00 | 0.00 | 7680.00 | 0.00 | 10656.00 | 0.00 | 10656.00 | 0.00 |
| 25 | Sales turnover [£] | 25778.83 | 712.76 | 29625.38 | 1284.20 | 13157.97 | 429.21 | 28683.18 | 1427.18 |
| 26 | Sales per employee [£] | 2577.88 | 71.28 | 2962.54 | 128.42 | 1096.50 | 35.77 | 2390.26 | 118.93 |
| 27 | Sales per employee hour worked [£] | 3.36 | 0.09 | 3.86 | 0.17 | 1.23 | 0.04 | 2.69 | 0.13 |
| 28 | Net profit per employee [£] | 386.68 | 10.69 | 444.38 | 19.26 | 164.47 | 5.37 | 358.54 | 17.84 |
| 29 | Net profit per hour worked [£] | 0.50 | 0.01 | 0.58 | 0.03 | 0.19 | 0.01 | 0.40 | 0.02 |

Table 6: Simulation output from testing proactive staff behaviour features in A&TV

| Measure | Parameter Settings | Womenswear | | | | | | | |
|---|---|---|---|---|---|---|---|---|---|
| | Scenario | a | | b | | c | | d | |
| | Staffing | optimised | | optimised | | real | | real | |
| | Number of {cashiers; normal staff; expert staff} | {3;8;2} | | {3;8;2} | | {2;13;1} | | {2;13;1} | |
| | Proactive | no | | yes | | no | | yes | |
| | **Non-monetary performance measures (means of weekly averages; times provided in minutes)** | | | | | | | | |
| | **Outputs** | **Mean** | **SD** | **Mean** | **SD** | **Mean** | **SD** | **Mean** | **SD** |
| 1 | Overall number of customers | 6384.95 | 10.80 | 6384.60 | 8.57 | 6381.63 | 13.22 | 6384.35 | 9.92 |
| 2 | Number of customers that leave happy (purchase) | 2931.43 | 2.01 | 3138.00 | 3.76 | 2004.97 | 1.31 | 2977.20 | 6.76 |
| 3 | % of customers that leave happy (purchase) | 45.91% | 0.07 | 49.15% | 0.06 | 31.42% | 0.07 | 46.63% | 0.09 |
| 4 | % of customers that leave not waiting for normal help | 0.00% | 0.00 | 0.00% | 0.00 | 0.00% | 0.00 | 0.00% | 0.00 |
| 5 | % of customers that leave not waiting for expert help | 0.03% | 0.00 | 0.03% | 0.00 | 0.09% | 0.01 | 0.10% | 0.01 |
| 6 | % of customers that leave not waiting to pay | 8.02% | 0.09 | 4.79% | 0.06 | 22.45% | 0.09 | 7.25% | 0.03 |
| 7 | % of customers left satisfied >0 (accumulated history) | 96.30% | 0.09 | 97.20% | 0.05 | 86.68% | 0.20 | 96.41% | 0.05 |
| 8 | % of customers left neutral =0 (accumulated history) | 2.34% | 0.04 | 2.11% | 0.03 | 3.79% | 0.07 | 2.35% | 0.03 |
| 9 | % of customers left unsatisfied <0 (accumulated history) | 1.36% | 0.07 | 0.69% | 0.04 | 9.53% | 0.16 | 1.24% | 0.04 |
| 10 | % of customers left satisfied >0 (experience per visit) | 49.42% | 0.07 | 52.07% | 0.07 | 37.60% | 0.08 | 49.99% | 0.09 |
| 11 | % of customers left neutral =0 (experience per visit) | 44.05% | 0.08 | 44.03% | 0.09 | 44.12% | 0.09 | 44.11% | 0.10 |
| 12 | % of customers left unsatisfied <0 (experience per visit) | 6.53% | 0.08 | 3.90% | 0.05 | 18.28% | 0.08 | 5.90% | 0.03 |
| 13 | Average time customers spend in department (per visit) | 23.19 | 0.02 | 23.04 | 0.02 | 24.11 | 0.02 | 23.62 | 0.02 |
| 14 | Average time customers spend waiting for help (per visit) | 2.37 | 0.12 | 2.32 | 0.14 | 3.58 | 0.09 | 3.60 | 0.10 |
| 15 | ... as % of time customers spend in help block | 17.51% | 0.88 | 17.17% | 1.05 | 26.38% | 0.66 | 26.54% | 0.71 |
| 16 | % of customers that have to wait for help | 0.62% | 0.04 | 0.71% | 0.05 | 1.20% | 0.05 | 1.26% | 0.06 |
| 17 | Average time customers spend waiting to pay (per visit) | 6.51 | 0.02 | 6.00 | 0.02 | 8.92 | 0.01 | 7.05 | 0.01 |
| 18 | ... as % of time customers spend in pay block | 71.73% | 0.08 | 68.37% | 0.08 | 82.80% | 0.05 | 71.49% | 0.03 |
| 19 | % of customers that have to wait to pay | 95.79% | 0.12 | 95.67% | 0.10 | 98.99% | 0.02 | 98.94% | 0.03 |
| 20 | Utilisation: cashiers | 93.07% | 0.06 | 92.98% | 0.05 | 95.47% | 0.04 | 95.29% | 0.05 |
| 21 | Utilisation: normal staff (excluding proactive support) | 27.67% | 0.15 | 30.02% | 0.11 | 19.53% | 0.09 | 26.63% | 0.12 |
| 22 | Utilisation: expert staff (excluding proactive support) | 26.44% | 0.61 | 27.26% | 0.51 | 20.69% | 0.27 | 21.86% | 0.28 |
| 23 | Utilisation: all staff (including proactive support) | 42.57% | 0.06 | 44.13% | 0.04 | 29.10% | 0.08 | 34.91% | 0.10 |
| | **Monetary performance measures (means of daily averages)** | | | | | | | | |
| | **Outputs** | **Mean** | **SD** | **Mean** | **SD** | **Mean** | **SD** | **Mean** | **SD** |
| 24 | Overall staff cost per day [£] | 12792.00 | 0.00 | 12792.00 | 0.00 | 18816.00 | 0.00 | 18816.00 | 0.00 |
| 25 | Sales turnover [£] | 17067.31 | 293.01 | 18262.05 | 586.83 | 11681.59 | 163.55 | 17330.39 | 745.67 |
| 26 | Sales per employee [£] | 1312.87 | 22.54 | 1404.77 | 45.14 | 730.10 | 10.22 | 1083.15 | 46.60 |
| 27 | Sales per employee hour worked [£] | 1.33 | 0.02 | 1.43 | 0.05 | 0.62 | 0.01 | 0.92 | 0.04 |
| 28 | Net profit per employee [£] | 525.15 | 9.02 | 561.91 | 18.06 | 292.04 | 4.09 | 433.26 | 18.64 |
| 29 | Net profit per hour worked [£] | 0.53 | 0.01 | 0.57 | 0.02 | 0.25 | 0.00 | 0.37 | 0.02 |

Table 7: Simulation output from testing proactive staff behaviour features in WW

There are some more interesting facts that can be observed from Table 6 and 7:

- Measure 6: The bottleneck in all scenarios for both department types seems to be the tills. Serving them proactively has reduced the problem but not eliminated it. This might be either due to the fact that we have misjudged the patience level of customers (the value we use is only an informed guess) or it is linked to the proactivity parameters we've used. However, even in the real system queues can get quite long and it takes a while until staff responds.
- Measure 13: The average time the customers spend in the department is neither affected by our consideration of proactivity nor by the different staffing levels we used. However, how they spend their time in the department is affected by both (see Measures 14-19). Most noteworthy is the observation that due to proactivity (normal service staff helping out as cashiers) in A&TV more customers have to queue for getting help, as normal service staff is busy at the tills and is not released immediately when required to help customers. However, overall the average queuing times in the help blocks are shorter when staff proactivity is modelled.
- Measure 20-23: All staff in A&TV is well utilised while normal service staff and expert staff in WW is having an extremely low utilisation in all scenarios. Although this is not directly obvious from Measure 21 and 22 (as these measures exclude proactive support) we can see from Measure 20 that the utilisation value of the cashiers has not changed and therefore the growth in overall staff utilisation (Measure 23) must be due to the proactive activities of normal service staff and expert staff. From our observation of the real system we can say that this is not the case. All WW staff is highly utilised, but some of the jobs they do besides serving at the till and responding to queries from customers (e.g. tidying up or replenishing shelves, manning changing rooms) we have not considered in our simulation model yet.
- Measure 29: The net profit per hour is very similar for the two department types, although different for the different scenarios. Unfortunately we don't have the figures from the case study company, to test if this is the same for the real system.

To summarise the results for this experiment we can say that by modelling proactiveness we are capturing the underlying mechanisms much better and therefore we do not have to tweak the staffing levels any more. Furthermore, our additional performance measures appear to be very useful for system analysis and support the analysis of cost factors for implementing different scenarios. However, in the analysis of the experiment we uncovered some problems with our staff utilisation capture, which we will have to investigate further.

*Experiment 2: Sensitivity Analysis*

Our second experiment is a calibration experiment and a sensitivity analysis in one. The main goal is to create a new set of proactivity parameter values that allows matching the average weekly number of transactions that we observed during our case study period (four weeks), i.e. 1787.09 for A&TV and 3172.35 for WW. The proactivity parameter values we are currently using are estimates and are the same for both departments. Besides delivering a new set of parameter values, the experiment will uncover, which of the proactivity parameters have the biggest impact on the behaviour of the simulation model.

Our research question for this experiment can be phrased as follows: *Can we calibrate our proactiveness parameter to match real system performance in terms of number of transactions?*

As we are already close to matching the true average weekly number of transactions value for WW with our current set of proactivity parameter values (see Table 5) we have limited the experiment to investigate only A&TV proactivity parameter values. Furthermore, due to time and resource constraints we will only test a subset of all possible parameters and their combinations.

We have developed the following strategy: For all scenarios we keep the overall number of staff members constant. First, we test the impact of the "critical queue length" parameter. We model an extreme case where we have all tills permanently manned (staffing level = {4;7;1}). This emulates a "critical queue length" of 0. Then we change this parameter in steps of 1 from 1 to 4, using real staffing levels. Second, we test the impact of the "maximum number of customers to serve as a temporary cashier" parameter. We change this parameter from 10 to 2 in steps of 2. Third, we test the impact of the three different strategies we implemented for closing down additional tills.

Table 8, 9 and 10 give a summary of the results of the experiment, presented in form of mean values and their standard deviations. We have not conducted a rigorous statistic analysis as we only want to demonstrate the new features here.

The impact test results for the "critical queue length" parameter are shown in Table 8. This parameter has the biggest impact but it is very difficult to estimate. The "critical queue length" that matches the "number of transactions" of the real system is approximately 1.25. This matches what we have observed in the A&TV department. Queues are kept very short and often customers are served directly at the till by the same person that has given the advice before (i.e. they don't have to queue). Therefore a "critical queue length" of 1.25 seems to be a plausible result. We also tested the impact of this parameter for WW where the "critical queue length" that matches the "number of transaction" of the real system is approximately 2.41.

| Parameter Settings | Audio & TV (outputs: means of weekly averages) | | | | | | | | | |
|---|---|---|---|---|---|---|---|---|---|---|
| Number of {cashiers; normal staff; expert staff} | {4;7;1} | | {1;10;1} | | {1;10;1} | | {1;10;1} | | {1;10;1} | |
| Proactive | no | | yes | | yes | | yes | | yes | |
| Critical queue length | - | | 1 | | 2 | | 3 | | 4 | |
| Outputs | Mean | SD | Mean | SD | Mean | SD | Mean | SD | Mean | SD |
| Number of customers that leave happy (purchase) | 1875.47 | 5.16 | 1833.18 | 6.99 | 1627.22 | 3.77 | 1348.48 | 2.85 | 1161.97 | 3.16 |
| % of customers that leave happy (purchase) | 45.87% | 0.15% | 44.84% | 0.17% | 39.85% | 0.07% | 32.99% | 0.09% | 28.42% | 0.08% |
| % of customers that leave not waiting for normal help | 2.35% | 0.04% | 1.08% | 0.05% | 0.75% | 0.03% | 0.46% | 0.02% | 0.32% | 0.02% |
| % of customers that leave not waiting for expert help | 1.86% | 0.04% | 1.95% | 0.03% | 1.92% | 0.03% | 1.89% | 0.04% | 1.88% | 0.03% |
| % of customers that leave not waiting to pay | 0.43% | 0.03% | 2.27% | 0.05% | 7.40% | 0.09% | 14.56% | 0.05% | 19.23% | 0.10% |

Table 8: Results from testing different "critical queue length" settings in A&TV

The impact test results for the "maximum number of customers to serve as a temporary cashier" parameter are shown in Table 9. For this test we have to use stop strategy 1: close down till opened by a normal service staff member after the "maximum number of customers to serve as a temporary cashier" have been served. We can see that with a decrease in the parameter value the number of customers that buy something is falling. When the parameter value is getting smaller (i.e. a temporary cashier serves less people in one go) it is becoming more likely that normal service staff that served as temporary cashier has to provide normal help before helping out again as a temporary cashier. Therefore, while the number of customers that leave not waiting to pay increases the number of customers that leave not waiting for normal service staff advice decreases. However, the impact is very small. We also tested the impact of this parameter for a "critical queue length" value of 2 and 1. We found that the smaller the "critical queue length" value the smaller the impact of the "maximum number of customers to serve as a temporary cashier" parameter on the "number of transactions".

| Parameter Settings | Audio & TV (outputs: means of weekly averages) | | | | | | | | | |
|---|---|---|---|---|---|---|---|---|---|---|
| Number of {cashiers; normal staff; expert staff} | {1;10;1} | | {1;10;1} | | {1;10;1} | | {1;10;1} | | {1;10;1} | |
| Proactive | yes | | yes | | yes | | yes | | yes | |
| Max. number of customers to serve as a temp. cashier | 10 | | 8 | | 6 | | 4 | | 2 | |
| Outputs | Mean | SD | Mean | SD | Mean | SD | Mean | SD | Mean | SD |
| Number of customers that leave happy (purchase) | 1502.98 | 5.30 | 1481.27 | 4.83 | 1451.80 | 0.97 | 1406.80 | 2.24 | 1336.83 | 2.93 |
| % of customers that leave happy (purchase) | 36.77% | 0.09% | 36.19% | 0.12% | 35.48% | 0.08% | 34.40% | 0.12% | 32.72% | 0.09% |
| % of customers that leave not waiting for normal help | 0.61% | 0.03% | 0.60% | 0.03% | 0.55% | 0.02% | 0.48% | 0.03% | 0.44% | 0.03% |
| % of customers that leave not waiting for expert help | 1.92% | 0.02% | 1.94% | 0.04% | 1.91% | 0.06% | 1.90% | 0.04% | 1.91% | 0.02% |
| % of customers that leave not waiting to pay | 10.63% | 0.08% | 11.09% | 0.09% | 11.91% | 0.08% | 13.01% | 0.09% | 14.78% | 0.07% |

Table 9: Results from testing different "max. number of customers served as temp. cashier" settings in A&TV

The impact test results for the "stop strategy" parameter are shown in Table 10. There are three different ways to determine when to close a temporary till: (1) after serving x people, where x is the maximum number of customers to be served by an individual temporary cashier in a row; (2) after queue length has been reduced below a critical value (critical queue length); (3) if either 1 or 2 is true. The table shows that stop strategy 1 results in the biggest turnover. Furthermore, it can be observed that stop strategy 2 and 3 produce nearly the same values for all performance measures listed. However, we also tested the impact of this parameter for a "critical queue length" value of 2 and 1 and found that this was just coincidence. In addition the test showed that the smaller the "critical queue length" value the smaller the impact of the "stop strategy" parameter on the "number of transactions". At a "critical queue length" of 1 there was no significant difference between values any more.

| Parameter Settings | Audio & TV (outputs: means of weekly averages) | | | | | |
|---|---|---|---|---|---|---|
| Number of {cashiers; normal staff; expert staff} | {1;10;1} | | {1;10;1} | | {1;10;1} | |
| Proactive | yes | | yes | | yes | |
| Stop strategy | 1 | | 2 | | 3 | |
| Outputs | Mean | SD | Mean | SD | Mean | SD |
| Number of customers that leave happy (purchase) | 1502.98 | 5.30 | 1350.17 | 3.08 | 1348.48 | 2.85 |
| % of customers that leave happy (purchase) | 36.77% | 0.09% | 33.00% | 0.07% | 32.99% | 0.09% |
| % of customers that leave not waiting for normal help | 0.61% | 0.03% | 0.48% | 0.03% | 0.46% | 0.02% |
| % of customers that leave not waiting for expert help | 1.92% | 0.02% | 1.90% | 0.05% | 1.89% | 0.04% |
| % of customers that leave not waiting to pay | 10.63% | 0.08% | 14.45% | 0.06% | 14.56% | 0.05% |

Table 10: Results from testing different "stop strategy" settings in A&TV

In summary we can say that the fact that the proactivity parameters are all somehow related to each other makes it a very difficult task to determine their values with confidence. Getting the "number of transactions" right does not mean that the individual parameters are right, this normally requires some insider knowledge. Also the variability and randomness in which proactive behaviour appears in the real system is very difficult to capture through simple rules and parameters. From the proactivity parameters tested we have identified the parameter "critical queue length" as having the biggest impact on simulation model output. More tests and observations are needed for developing a more reliable set of proactivity rules and parameter values.

## Conclusions

In this paper we have investigated how different levels of staff proactiveness impact on the performance output of our department store simulation model. This investigation was possible as we used a new modelling approach - a combined DES/ABS approach. The system flow (in our case queues) is modelled using a DES queuing model and the staff and customers are modelled using the agent paradigm. This paradigm allows considering autonomy and proactiveness, two attributes vital for more realistic modelling of human centric complex adaptive systems.

Overall, our experiments have shown that proactive behaviour is a very important component for modelling the true dynamics that are inherent in people centric retail system such as department stores. Since we implemented proactive staff behaviour (in the form of proactive till manning) simulation results have been much closer to what we observed in the real system with regards to system dynamics and system performance.

However, the utilisation result for WW from Experiment 1 revealed that we are still not capturing all the relevant (high impact) system dynamics. More work is needed to identify the additional tasks that need to be considered. If these tasks are not directly relevant to our investigation we only need to know the amount of time staff is typically spends on them. Then we can model them as a collective task that either can be interrupted (proactively after a short period of contemplating or directed by a section manager) or which cannot be interrupted (working in the back sorting or ordering stock). It is important to consider these tasks as otherwise we will get inaccurate utilisation values and staff would be always available to help out proactively, which as we can see in WW gives the wrong impression.

From Experiment 2, we can learn that the adjustment of proactivity parameters and rules is possible, but difficult. In some regards, this is not different from other systems with many variables; however, there is added difficulty due to the relatively high correlation of these measures. We have shown that careful manipulation of these parameters allows us to calibrate the system further. However, our experiments also reveal a shortcoming in our approach. The way we have implemented proactivity does not yet fully cover the variability that occurs in real life. For instance, not all sales staff will respond equally to the same "critical queue length", some will have more and others less tolerance to letting customers wait. A potential future area of work is to try and make these measures more robust by coupling them to the staff agents' attributes.

Overall, our new performance measures turned out to be very useful for an analysis of the system: on the one hand to gain better insight into the dynamics of the system (utilisation measures) and on the other hand for optimising the system (utilisation and productivity / profitability measures).

In the future, we are interested in exploring other ways of implementing the agent decision making process. It has been argued that modelling the autonomous internal decision making logic of customers is a crucial element for simulation models of consumer behaviour (Jager 2007). It would be interesting to compare such an approach to the probabilistic one we have currently in place, in particular as we have not found any such study in the literature. Furthermore, it still needs to be tested how our simulation models prove themselves in practice. As yet they have not been applied in a real case of decision making by the participating case study organisation. Once we have passed this milestone the true value of our efforts can be evaluated. If the proposed strategic changes will be implemented in the real system it would be possible to validate the implemented solution against the model results.

In conclusion, we can say that our choice of using a combined DES/ABS approach for modelling the impact of people management practices has been a good choice as we were able to improve the accuracy of system representation. Moreover, the resultant systems take the best from both worlds and relatively compact simulation models allow for realistic scenario analysis. Finally, we can suggest to practitioners with a clear conscience that when modelling service systems with a high proportion of human interaction the extra work of adding proactivity to the simulation model is worth the effort, i.e. it does produce a significant improvement of the simulation output accuracy that justifies the extra costs for data collection and modelling.

## Acknowledgement

The research presented in this paper has been conducted in collaboration with Prof. Chris Clegg and Helen Celia from the University of Leeds, Centre for Organisational Strategy, Learning & Change (LUBS), Leeds, LS2 9JT, UK. The research is funded by the UK Engineering and Physical Sciences Research Council under Grand Number EPSRC EP/D503949/1

## References

- AJZEN I (1985) From intentions to actions: A theory of planned behavior. In Kuhl J and Beckmann J (Eds.) *Action-Control: From Cognition to Behavior*, Heidelberg, Germany: Springer, pp. 11-39.

- BAYDAR C (2003) Agent-based modelling and simulation of store performance for personalised pricing. In Chich S, Sanchez P J, Ferrin D, and Morrice D J (Eds) *Proceedings of the 35th Winter Simulation Conference, New Orleans, Louisiana, USA, December 7-10*, pp. 1759-1764.

- BAXTER N, Collings D and Adjali I (2003) Agent-based modelling - Intelligent customer relationship management. *BT Technology Journal*, 21(2): 126-132.

- BENNETT R (1998) Queues, customer characteristics and policies for managing waiting-lines in supermarkets. *International Journal of Retail and Distribution Management*, 26(2): 78-87.

- BERMAN O and Larson R C (2004) A queuing control model for retail services having back room operations and cross-trained workers. *Computers & Operations Research*, 31: 201-222.

- BIRDI K, Clegg C, Patterson M, Robinson A, Stride C B, Wall T and Wood S J (2008) The impact of human resource and operational management practices on company productivity: A longitudinal study. *Personnel Psychology*, 61: 467-501.

- BLOOM N, Dorgan S, Dowdy J, Rippin T and Van Reenen J (2005) Management practices: The impact on company performance. *Centrepiece*, 10(2): 2-6.

- BOERO R and Squazzoni F (2005) Does empirical embeddedness matter? Methodological issues on agent-based models for analytical science. *Journal of Artificial Societies and Social Simulation*, 8(4) 6 http://jasss.soc.surrey.ac.uk/8/4/6.html.

- BONABEAU E (2002) Agent-based modeling: Methods and techniques for simulating human systems. *Proceedings of the National Academy of Science of the USA*, 99(3): 7280-7287.

- CAO J (1999) Evaluation of advertising effectiveness using agent-based modeling and simulation. In *Proceedings of 2nd UK Workshop of SIG on Multi-Agent Systems. Bristol, UK, December 6-7*.

- CASTI J (1999) Firm forecast. *NewScientist*, 2183: 42-46.

- CONSIGLIO A (2007). *Artificial Markets Modeling: Methods and Applications*. Berlin, Germany: Springer.

- CRANT J M (2000) Proactive behavior in organizations. *Journal of Management*, 26: 435–462.

- CRESPI G, Criscuolo C, Haskel J and Hawkes D (2006) Measuring and understanding productivity in UK market services. *Oxford Review of Economic Policy*, 22(4): 560-572.

- CSIK B (2003) Simulation of competitive market situations using intelligent agents. *Periodica Polytechnica Social and Management Sciences*, 11: 83-93.

- DARLEY V, von Tessin P and Sanders D (2004) An agent-based model of a corrugated-box factory: The trade-off between finished goods stock and on-time-in-full delivery. In *Proceedings of the 5th Workshop on Agent-Based Simulation. Lisbon, Portugal, May 3-5*, pp. 113-118.

- DUBIEL B and Tsimhoni O (2005) Integrating agent based modeling into a discrete event simulation. In Kuhl M E, Steiger N M, Armstrong F B and Joines J A (Eds.) *Proceedings of the 37th Winter Simulation Conference, Orlando, FL, USA, December 4-7*, pp.1029-1037.

- FRESE M and Fay D (2001) Personal initiative: An active performance concept for work in the 21st century. *Research in Organizational Behavior*, 23: 133–188.

- GILBERT N and Terna P (2000) How to build and use agent-based models in social science. *Mind & Society*, 1(1): 57-72.

- GILBERT N and Troitzsch K G (2005) *Simulation for the Social Scientist, 2nd Ed.* Buckingham, UK: Open University Press.

- GRIFFITH R and Harmgart H (2005) Retail productivity. *Advanced Institute of Management Research (AIM) Working Paper Series 017*, ISSN 1744-0009.

- HANSEN T (2005) Perspectives on consumer decision making: An integrated approach. *Journal of Consumer Behaviour*, 4(6): 420-437.

- HOLLOCKS B (2006) Forty years of discrete-event simulation - a personal reflection. *Journal of the Operational Research Society*, 57(12): 1383-1399.

- HUDSON R (2000) Book Review: Introducing human geographies. *Progress in Human Geography*, 24: 677

- JAGER W (2007) The four P's in social simulation, a perspective on how marketing could benefit from the use of social simulation. *Journal of Business Research*, 60: 868–875.

- JAGER W, Janssen M A, De Vries H J M, De Greef J and Vlek C A J (2000) Behaviour in commons dilemmas: Homo Economicus and Homo Psychologicus in an ecological-economic model. *Ecological Economics*, 35(3): 357-379.

- JANSSEN M A and Jager W (2001) Fashions, habits and changing preferences: Simulation of psychological factors affecting market dynamics. Journal of Economic Psychology, 22(6): 745-772.

- KELLY J P, Smith S M and Hunt H K (2000) Fulfilment of planned and unplanned purchases of sale- and regular-price items: A benchmark study. *International Review of Retail, Distribution and Consumer Research*, 10(3): 247-263.

- KITAZAWA K and Batty M (2004) Pedestrian behaviour modelling: An application to retail movements using a genetic algorithm. In *Proceedings of the 7th International Conference on Design and Decision Support Systems in Architecture and Urban Planning. St Michelsgestel, Netherlands, July 2-5.*

- KORITAROV V (2004) Real-world market representation with agents: Modeling the electricity market as a complex adaptive system with an agent-based approach. *IEEE Power and Energy Magazine*, July/August 2004: 39-46.

- LARSSON J and Petersson J (2007) Agent-based modelling in fashion demand chains. In Halldórsson Á and Stefánsson G (Eds.) *Proceedings of the 19th Annual NOFOMA Conference, Reykjavik, Iceland, June 7-8.*

- MOSS S and Edmonds B (2005) Sociology and simulation: Statistical and qualitative cross-validation. *American Journal of Sociology*, 110: 1095-1131.

- NICHOLSON M, Clarke I and Blakemore M (2002) "One brand, three ways to shop": Situational variables and multichannel consumer behaviour. *International Review of Retail, Distribution and Consumer Research*, 12: 131-148.

- NORTH M J and Macal C M (2007) *Managing Business Complexity: Discovering Strategic Solutions with Agent-Based Modeling and Simulation*. New York, USA: Oxford University Press.

- PARASURAMAN A, Zeithaml V A and Berry L L (1988) SERVQUAL: A Multi-item scale measuring consumer perceptions of service quality. *Journal of Retailing*, 64(1): 12-37.

- PARUNAK H V D, Savit H R and Riolo R L. Agent-based modeling vs. equation-based modeling: A case study and users' guide. In Sichman J S, Conte R and Gilbert N (Eds.) *Multi-Agent Systems and Agent-Based Simulation, Lecture Notes in Artificial Intelligence (LNAI) Vol. 1534*. Berlin, Germany: Springer, pp. 10-25.

- PATEL S and Schlijper A (2004) *Models of Consumer Behaviour*. Smith Institute (Unilever), UK.

- PATTERSON M, West M A, Lawthom R and Nickell S (1997) *Impact of People Management Practices on Business Performance*. London, UK: Institute of Personnel and Development.

- POURDEHNAD J, Maani K and Sedehi H (2002) System dynamics and intelligent agent-based simulation: Where is the synergy? In Davidsen P I, Mollona E, Diker V G, Langer R S and Rowe J I (Eds.) *Proceedings of the 20th International Conference of the System Dynamics Society, Palermo, Italy, July 28 - August 1*, p. 118.

- RANK J, Carsten J M, Unger J M and Spector P E (2007) Proactive customer service performance: Relationships with individual, task, and leadership variables. *Human Performance*, 20: 363-390.

- RAO A S and Georgeff M P (1995) BDI agents: From theory to practice. In Lesser V R and Gasser L (Eds.) *Proceedings of the 1st International Conference on Multi-Agent Systems. San Francisco, CA, USA, June 12-14*, pp. 312-319.

- REYNOLDS K E and Beatty S E (1999) A relationship customer typology. *Journal of Retailing*, 75(4): 509-523.

- ROBINSON S (2004) *Simulation: The Practice of Model Development and Use*. Chichester, UK: Wiley.

- RUDMAN R (2008) People management and the bottom line. In *PsychPress Newsletter 12/2008*, http://www.psychpress.com.au/psychometric/files/newsletter_dec08_pplmgt.pdf?&topic=300.

- SAID L B and Bouron T (2001) Multi-agent simulation of consumer behaviours in a competitive market. In Lund H H, Mayoh B H and Perram J W (Eds.) *Proceedings of the 10th European Workshop on Multi-Agent Systems, Modelling Autonomous Agents in A Multi-Agent Word. Annecy, France, May 2-4*, pp. 31-43.

- SAID L B, Bouron T and Drogoul A (2002) Agent-based interaction analysis of consumer behaviour. In *Proceedings of the First International Joint Conference on Autonomous Agents and Multiagent Systems. Bologna, Italy, July 15-19*, pp. 184-190.

- SCHENK T A, Loeffler G and Rau J (2007) Agent-based simulation of consumer behaviour in grocery shopping on a regional level. *Journal of Business Research*, 60(8): 894-903.

- SCHWAIGER A and Stahmer B (2003) SimMarket: Multi-agent based customer simulation and decision support for category management. In Schillo M, Klusch M, Muller J and Tianfield H (Eds.) Multiagent System Technologies, Lecture Notes in Artificial Intelligence (LNAI) Vol. 2831, Berlin, Germany: Springer, pp. 74-84.

- SELLERS-RUBIO R and Mas-Ruiz F (2007) Different approaches to the evaluation of performance in retailing. *The International Review of Retail, Distribution and Consumer Research*, 17(5): 503-522.

- SHOPSIM (2010) http://www.savannah-simulations.com/.

- SIEBERS P O (2010) http://www.openabm.org/ (we have provided an applet version of the model and the source code; a copy of AnyLogic v5.5 is required to view and alter the source code).

- SIEBERS P O, Aickelin U, Battisti G, Celia H, Clegg C W, Fu X, De Hoyos R, Iona A, Petrescu A and Peixoto, A (2008) The role of management practices in closing the productivity gap. *Advanced Institute of Management Research (AIM) Working Paper Series 065*, ISSN 1744-0009.

- SIEBERS P O, Aickelin U, Celia H and Clegg C (2009) Modeling and simulating retail management practices: A first approach. *International Journal of Simulation and Process Modelling*, 5(3): 215-232.

- SIEBERS P O, Aickelin U, Celia H and Clegg C (2010a) Towards the development of a simulator for investigating the impact of people management practices on retail performance. *Journal of Simulation*, in print. Available online: http://www.palgrave-journals.com/jos/journal/vaop/ncurrent/abs/jos201020a.html

- SIEBERS P O, Macal C M, Garnett J, Buxton D and Pidd M (2010b) Discrete-event simulation is dead, long live agent-based simulation. *Journal of Simulation*, 4(3): 204-210.

- SIMMONS M (1989) Customer attitudes toward service. *International Journal of Retail and Distribution Management*, 17(6): 6-8.

- SIMON F and Usunier J C (2007) Cognitive, demographic and situational determinants of service customer preference for personnel-in-contact over self-service technology. *International Journal of Research in Marketing*, 24(2): 163-173.

- TAI S H C and Fung A M C (1997) Application of an environmental psychology model to in-store buying behaviour. *International Review of Retail, Distribution and Consumer Research*, 7(4): 311-337.

- TESFATSION L and Judd K L (2006) *Handbook of Computational Economics: Agent-Based Computational Economics, Vol. 2*. Oxford, UK: Elsevier.

- TWOMEY P and Cadman R (2002) Agent-based modeling of customer behavior in the telecoms and media markets. *Info - The Journal of Policy, Regulation and Strategy for Telecommunications*, 4(1): 56-63.

- VAN RIEMSDIJK B, Dastani M and Winikoff M (2008) Goals in agent systems: A unifying framework. In Padgham L, Parkes DC, Müller J P and Parsons S (Eds.) *Proceedings of the 7th International Joint Conference on Autonomous Agents and Multiagent Systems. Estoril, Portugal, May 12-16*, pp. 713-720.

- VRIEND N J (1995) Self-organization of markets: An example of a computational approach. *Computational Economics*, 8: 205-231.

- WINIKOFF M, Padgham L, Harland J and Thangarajah J (2002) Declarative and procedural goals in intelligent agent systems. In Fensel D, Giunchiglia F, McGuinness D L and Williams M A (Eds.) *Proceedings of the 8th International Conference on Principles of Knowledge Representation and Reasoning. Toulouse, France, April 22-25*, pp. 470-481.

- XJTEK (2005) *AnyLogic User's Guide*. St. Petersburg, Russian Federation: XJ Technologies Company.

- YOUN S and Faber R J (2000) Impulse buying: Its relation to personality traits and cues. *Advances in Consumer Research*, 27(1): 179-185.